\documentclass[11pt,a4paper]{article}
\usepackage[hyperref]{ranlp2023}
\usepackage{times}
\usepackage{latexsym}

\usepackage{microtype}

\usepackage{multirow}
\usepackage[nolist]{acronym}
\usepackage{subcaption}
\usepackage{graphicx}
\usepackage[mathscr]{euscript}
\DeclareSymbolFont{rsfs}{U}{rsfs}{m}{n}
\DeclareSymbolFontAlphabet{\mathscrsfs}{rsfs}
\usepackage{algorithm}
\usepackage[noend]{algpseudocode}
\usepackage{adjustbox,booktabs}
\usepackage{dblfloatfix} 
\usepackage{booktabs}

\aclfinalcopy 

\title{Efficient Domain Adaptation of Sentence Embeddings Using Adapters}

\author{Tim Schopf, Dennis N. Schneider, and Florian Matthes \\
         Technical University of Munich, Department of Computer Science, Germany \\
         \texttt{\{tim.schopf,dennis.schneider,matthes\}@tum.de}}

\date{}

\begin{acronym}
\acro{nlp}[NLP]{natural language processing}
\acro{nli}[NLI]{natural language inference}
\acro{lm}[LM]{Language Model}
\acro{plm}[PLM]{Pretrained Language Model}
\acroplural{plm}[PLMs]{Pretrained Language Models}
\acro{sts}[STS]{semantic textual similarity}
\acro{kg}[KG]{knowledge graph}
\acroplural{kg}[KGs]{knowledge graphs}
\acro{pwc}[PwC]{Papers with Code}
\acro{p}[P]{Precision}
\acro{r}[R]{Recall}
\acro{mrr}[MRR]{Mean Reciprocal Rank}
\acro{map}[MAP]{Mean Average Precision}
\end{acronym}

\newcommand{\argmin}{\mathop{\mathrm{argmin}}\limits} 

\begin{document}
\maketitle
\begin{abstract}

Sentence embeddings enable us to capture the semantic similarity of short texts. Most sentence embedding models are trained for general semantic textual similarity tasks. Therefore, to use sentence embeddings in a particular domain, the model must be adapted to it in order to achieve good results. Usually, this is done by fine-tuning the entire sentence embedding model for the domain of interest. While this approach yields state-of-the-art results, all of the model's weights are updated during fine-tuning, making this method resource-intensive. Therefore, instead of fine-tuning entire sentence embedding models for each target domain individually, we propose to train lightweight adapters. These domain-specific adapters do not require fine-tuning all underlying sentence embedding model parameters. Instead, we only train a small number of additional parameters while keeping the weights of the underlying sentence embedding model fixed. Training domain-specific adapters allows always using the same base model and only exchanging the domain-specific adapters to adapt sentence embeddings to a specific domain. We show that using adapters for parameter-efficient domain adaptation of sentence embeddings yields competitive performance within 1\% of a domain-adapted, entirely fine-tuned sentence embedding model while only training approximately 3.6\% of the parameters.

\end{abstract}

\section{Introduction}
\label{sec:introduction}

Learning sentence embeddings is an essential task in \ac{nlp} and has already been extensively investigated in the literature \cite{NIPS2015_f442d33f,hill-etal-2016-learning,conneau-etal-2017-supervised,logeswaran2018an,cer-etal-2018-universal,reimers-gurevych-2019-sentence,gao-etal-2021-simcse,wu-etal-2022-esimcse,schopf2023aspectcse,schopf2023exploring}. Sentence embeddings are especially useful in information retrieval \cite{NEURIPS2020_6b493230,schopf_etal_kdir_22,schneider-etal-2022-decade} or unsupervised text classification settings \cite{schopf_etal_webist_21,10.1145/3582768.3582795,10.1007/978-3-031-24197-0_4}. Lately, the most popular approach for sentence embedding learning is to fine-tune pretrained language models with a contrastive learning objective \cite{liu-etal-2021-dialoguecse,zhang-etal-2022-mcse,chuang-etal-2022-diffcse,nishikawa-etal-2022-ease,cao-etal-2022-exploring,jiang-etal-2022-improved}. While this approach provides state-of-the-art results, all of the model's weights are updated during fine-tuning, making this method resource-intensive. This is a problem, particularly when domain-specific models are needed. Then, a specialized model must be trained for each domain of interest, resulting in resource-intensive training. 

\begin{figure}[t!]
    \centering
    \includegraphics[width=0.8\columnwidth]{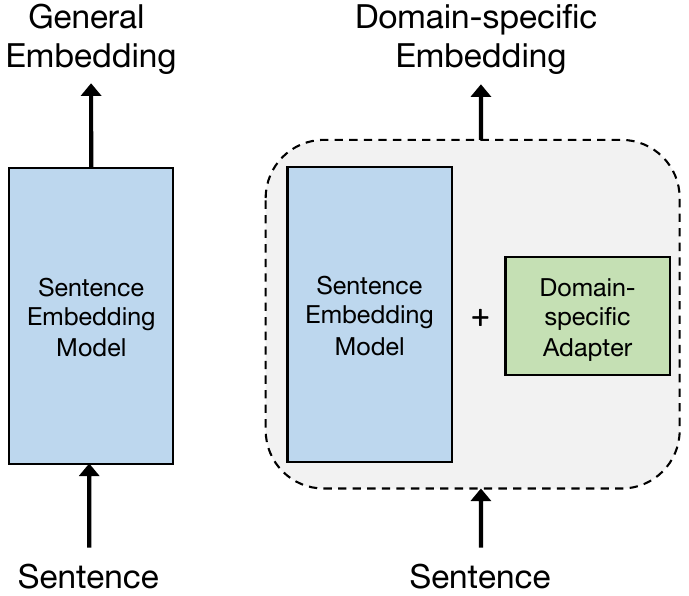}
    \caption{Sentence embedding models are usually trained to obtain state-of-the-art sentence representations for general semantic textual similarity tasks. By injecting domain-specific knowledge of adapters into the sentence embedding model, we can efficiently adapt the resulting representations for semantic textual similarity tasks in different domains.}
    \label{fig:example-visualization}
\end{figure}

Recently, \textit{adapters} have emerged as a parameter-efficient strategy to fine-tune \acp{lm}. Adapters do not require fine-tuning of all parameters of the pretrained model and instead introduce a small number of task-specific parameters while keeping the underlying pretrained language model fixed \cite{pfeiffer-etal-2021-adapterfusion}. They enable efficient parameter sharing between tasks and domains by training many task-specific, domain-specific, and language-specific adapters for the same model, which can be exchanged and combined post-hoc \cite{pfeiffer-etal-2020-adapterhub}. Therefore, many different adapter architectures have been proposed for various domains and tasks \cite{pfeiffer-etal-2020-mad,pfeiffer-etal-2021-unks,DBLP:journals/corr/abs-2012-06460,he-etal-2021-effectiveness,le-etal-2021-lightweight,parovic-etal-2022-bad,lee-etal-2022-fad}. However, to the best of our knowledge, no method currently exists for efficient domain adaptation of sentence embeddings using adapters.

In this paper, we aim to bridge this gap by proposing approaches for adapter-based domain adaptation of sentence embeddings, allowing us to train models for many different domains efficiently. Therefore, we investigate how to adapt general pretrained sentence embedding models to different domains using domain-specific adapters. As shown in Figure \ref{fig:example-visualization}, this allows always using the same base model to adapt sentence embeddings to a specific domain and only needing to exchange the domain-specific adapters. Accordingly, we train lightweight adapters for each domain and avoid expensive training of entire sentence embedding models.

\section{Related Work}
\label{sec:related-work}

Adapters have been introduced by \citet{pmlr-v97-houlsby19a} as a parameter-efficient alternative for task-specific fine-tuning of language models. Since their introduction, adapters have been used to fine-tune models for single tasks as well as in multi-task settings \cite{pfeiffer-etal-2021-adapterfusion}. Usually, adapters are used to solve tasks such as classification \cite{lauscher-etal-2020-common}, machine translation \cite{baziotis-etal-2022-multilingual}, question answering \cite{pfeiffer-etal-2022-xgqa}, or reasoning \cite{pfeiffer-etal-2021-adapterfusion}. While there exist adapters for \acf{sts} tasks on the \textit{AdapterHub} \cite{pfeiffer-etal-2020-adapterhub}, these are trained on general \ac{sts} datasets using a task-unspecific pretrained language model as a basis. We, however, focus on adapting pretrained sentence embedding models to specific domains using adapters.

\section{Method}
\label{sec:method}

We assume we have a base sentence embedding model from the source domain and labeled datasets for each target domain. Instead of fine-tuning the entire sentence embedding model for each target domain individually, we train lightweight adapters for each domain. This domain-specific fine-tuning with adapters involves adding a small number of new parameters to the sentence embedding model. During training, the parameters of the sentence embedding model are frozen, and only the weights of the adapters are updated. Formally, we adopt the general definition for adapter-based fine-tuning of \citet{pfeiffer-etal-2021-adapterfusion} as follows: 

For each of the $N$ domains, the sentence embedding model is initialized with parameters $\Theta_{0}$. Additionally, a set of new and randomly initialized adapter parameters $\Phi_{n}$ are introduced. The parameters $\Theta_{0}$ are fixed and only the parameters $\Phi_{n}$ are trained. Given training data $D_{n}$ and a loss function $L$, the objective for each domain $n \in {1,...,N}$ is of the form:

\begin{equation}
\label{eqn:adapter_eqn}
    \Phi_{n} \leftarrow \argmin_{\Phi} L(D_{n};\Theta_{0},\Phi)
\end{equation}

Usually, the adapter parameters $\Phi_{n}$ are significantly less than the parameters $\Theta_{0}$ of the base model \cite{pfeiffer-etal-2021-adapterfusion}, e.g., only 3.6\% of the parameters of the pretrained model in \citet{pmlr-v97-houlsby19a}.

\section{Experiments}
\label{sec:experiments}

In this section, we describe the used adapter architectures, loss functions, and datasets. In all experiments, we use $\textrm{SimCSE}_{sup-bert-base}$ \cite{gao-etal-2021-simcse} as the base sentence embedding model. It is trained on \ac{nli} datasets \cite{bowman-etal-2015-large,williams-etal-2018-broad} for \ac{sts} tasks in the general domain. We fine-tune all models and adapters for five epochs using a learning rate of $1e^{-5}$.

\subsection{Adapter Architectures}
\label{subsec:adapter-architectures}

We investigate how different adapter architectures affect the domain adaptability of sentence embedding models.

\paragraph{Houlsby-Adapter} This adapter, introduced by \citet{pmlr-v97-houlsby19a}, uses a bottleneck architecture. The adapter modules are added after both the multi-head attention and feed-forward block in each transformer layer \cite{NIPS2017_3f5ee243} of the base model. The adapter layers transform their input into a very low-dimensional representation and upsample it again to the same dimension in the output. This generates a parameter-efficient lower-dimensional representation while most information is kept.

\begin{figure}[H]
    \centering
    \includegraphics[width=0.8\columnwidth]{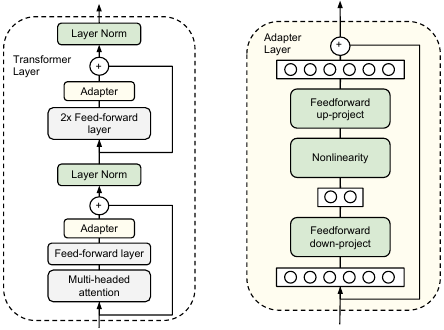}
    \caption{Houlsby-Adapter architecture as introduced by \citet{pmlr-v97-houlsby19a}. On the left side, the adapter is illustrated to be added twice to each transformer layer. Once after the multi-head attention and once after the feed-forward layer. On the right side, the bottleneck architecture of the adapter is presented.}
    \label{fig:houlsby-adapter}
\end{figure}

\paragraph{Pfeiffer-Adapter}  This adapter, introduced by \citet{pfeiffer-etal-2021-adapterfusion}, also uses a bottleneck architecture. However, the adapter modules are added only after the feed-forward block in each transformer layer of the base model. This architecture allows merging multiple adapters trained on different tasks. In this work, however, this multitask learning capability is not needed, and we only use the single-task mode.

\begin{figure}[H]
    \centering
    \includegraphics[width=0.35\columnwidth]{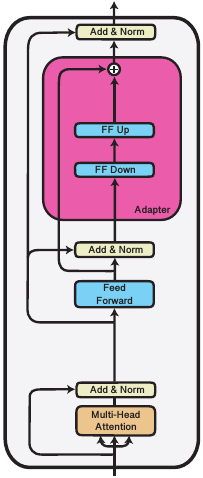}
    \caption{Pfeiffer-Adapter architecture as introduced by \citet{pfeiffer-etal-2021-adapterfusion}. Unlike the Houlsby-Adapter, a single Pfeiffer-Adapter is added in each transformer block only after the forward layer.}
    \label{fig:pfeiffer-adapter}
\end{figure}

\paragraph{K-Adapter} This adapter, introduced by \citet{wang-etal-2021-k}, works as outside plug-in for the base model. Each adapter model consists of $K$ adapter layers containing $N$ transformer layers and two projection layers across which a skip connection is applied. The adapter layers combine the output of an intermediate transformer layer in the base model with the output of a previous adapter layer. To generate the final output, the last hidden states of the adapter are concatenated with the last hidden states of the base model and transformed into the correct output dimension with a simple dense layer.

\begin{figure}[H]
    \centering
    \includegraphics[width=0.8\columnwidth]{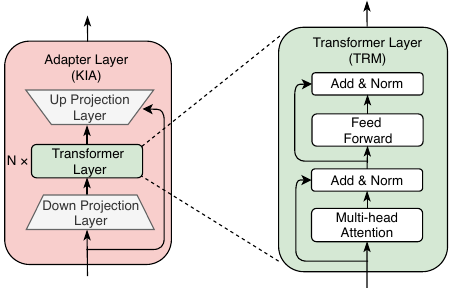}
    \caption{K-Adapter architecture as introduced by \citet{wang-etal-2021-k}. The adapter layer (left) consists of two projection layers, $N=2$ transformer layers, and a skip connection between two projection layers. The adapter layers are plugged among different transformer layers of the base model. The final output consists of the concatenated last hidden states of the adapter and the base model.}
    \label{fig:k-adapter}
\end{figure}

For reference, Table \ref{tab:adapter-parameter} shows the number of parameters per adapter model compared to commonly used base models, highlighting the efficient nature of adapters.

\begin{table}[H]
\centering
\resizebox{\columnwidth}{!}{%
\begin{tabular}{|l|c|c|}
\hline
& \textbf{BERT-base} & \textbf{RoBERTa-large} \\ \hline
\begin{tabular}[c]{@{}l@{}}No. of Parameters \\ \textbf{Base Model}\end{tabular}      & 110M               & 355M                   \\ \hline
\begin{tabular}[c]{@{}l@{}}No. of Parameters \\ \textbf{Houlsby-Adapter}\end{tabular}  & 4M                 & 6M                     \\ \hline
\begin{tabular}[c]{@{}l@{}}No. of Parameters \\ \textbf{Pfeiffer-Adapter}\end{tabular} & 10M                & 12M                    \\ \hline
\begin{tabular}[c]{@{}l@{}}No. of Parameters \\ \textbf{K-Adapter}\end{tabular}       & 47M                & 47M                    \\ \hline
\end{tabular}%
}
\caption{Number of trainable Parameters for different base models and adapter architectures.}
\label{tab:adapter-parameter}
\end{table}

\begin{table*}[t]
    \centering
    \resizebox{1.94\columnwidth}{!}{%
    \renewcommand{\arraystretch}{1.2} %
   
    \begin{adjustbox}{center}

    \begin{tabular}{cl|c|cccc|c}
    \toprule
    \multicolumn{2}{l}{\textbf{Datasets $\rightarrow$}} & \multicolumn{1}{|c|}{\textbf{AskUbuntu}} & \multicolumn{4}{|c|}{\textbf{SciDocs}} & \multicolumn{1}{|c}{\textbf{Average}} \\
    

    \multicolumn{2}{l|}{\textbf{Models $\downarrow$}} &  & \textbf{Cite} & \textbf{CC} & \textbf{CR} & \textbf{CV} & \\
    
    \midrule
    \multicolumn{2}{l|}{\textit{Out-of-the-box} SimCSE \textit{(lower bound)}} & 60.3 & 79.3 & 82.10 & 76.87 & 78.36 & 75.39 \\ \hline
    \multirow{4}{*}{\rotatebox[origin=c]{0}{\parbox[c]{0.1cm}{\centering $\ell_{1}$ \newline}}} & \multicolumn{1}{l|}{Houlsby-Adapter} & \underline{64.0} & \underline{\textbf{88.2}} & 88.69 & \underline{\textbf{82.42}} & \underline{83.99} & 81.46 \\
    & \multicolumn{1}{l|}{Pfeiffer-Adapter} & 63.8 & 87.8 &\underline{88.73} & 81.65 & 83.27 & 81.05 \\
    & \multicolumn{1}{l|}{K-Adapter} & 62.5 & 85.6 & 87.70 & 80.09 & 82.85 & 79.75 \\ 

    \cmidrule(lr){2-8}
    
    & \multicolumn{1}{l|}{\textit{In-domain supervised} SimCSE \textit{(upper bound)}} & 65.3 & 88.0 & 87.74 & 84.15 & 83.32 & 81.70\\
    \hline
    \multirow{4}{*}{\rotatebox[origin=c]{0}{\parbox[c]{0.1cm}{\centering $\ell_{2}$}}}
     & \multicolumn{1}{l|}{Houlsby-Adapter} & \underline{\textbf{64.5}} & \underline{87.3} & \underline{\textbf{89.01}} & \underline{82.41} & \underline{\textbf{84.42}} & \underline{\textbf{81.53}} \\
    {} & \multicolumn{1}{l|}{Pfeiffer-Adapter} & 64.2 & 87.0 & 88.63 & 81.98 & 84.41 & 81.24 \\
    & \multicolumn{1}{l|}{K-Adapter} & 62.8 & 85.3 & 87.92 & 80.05 & 83.29 & 79.87 \\

    \cmidrule(lr){2-8}
    
    & \multicolumn{1}{l|}{\textit{In-domain supervised} SimCSE \textit{(upper bound)}} & 65.2 & 88.3 & 88.11 & 84.46 & 83.63 & 81.94 \\
    \bottomrule
    \end{tabular}
\end{adjustbox}%
}
\caption{Evaluation results of the adapter-based domain adaptation using the different loss functions $\ell_{1}$ and $\ell_{2}$. The evaluation metric is \ac{map}. We show the performance of the SimCSE model without domain-specific fine-tuning as a lower bound. Additionally, we show the performance of SimCSE models using traditional fine-tuning with the respective loss functions as upper bounds. For the upper bounds, all model weights have been updated during training. In contrast, only the adapter weights were updated during adapter training while the base model parameters were frozen. In bold, we highlight the best adapter performance overall and underline the best adapter results per loss function.}
\label{tab:evaluation}
\end{table*}

\subsection{Loss Functions}
\label{subsec:loss-functions}

We investigate two different loss functions that are proven to teach models to learn a notion of \ac{sts} from triplets of examples. We assume a set of triplets $\mathcal{D} = \{(x_{i},x_{i}^{+},x_{i}^{-})\}$, where $x_{i}$ is an anchor sentence, $x_{i}^{+}$ is a positive sample and $x_{i}^{-}$ is a negative sample. With $h_{i}$, $h_{i}^{+}$, and $h_{i}^{-}$ as representations of $x_{i}$, $x_{i}^{+}$, and $x_{i}^{-}$, we use the triplet margin loss function of \citet{cohan-etal-2020-specter} as follows:

\begin{equation}\label{eq:specter-loss}
\resizebox{0.89\linewidth}{!}{$
  \ell_{1} = \max \{(d(h_{i},h_{i}^{+}) - d(h_{i},h_{i}^{-}) + m),0\}$}
\end{equation}

where $d$ is the L2 norm distance function and $m$ is the loss margin hyperparameter set to 1. 

Additionally, we use the contrastive objective of \citet{gao-etal-2021-simcse} as follows: 

\begin{equation}\label{eq:simcse-loss}
\resizebox{.89\linewidth}{!}{$
  \ell_{2} = -\log \frac{e^{sim(h_{i},h_{i}^{+})/\tau}}{\sum_{j=1}^{N}(e^{sim(h_{i},h_{j}^{+})/\tau} + e^{sim(h_{i},h_{j}^{-})/\tau})}$}
\end{equation}

with a mini-batch of $N$ triplets, a temperature hyperparameter $\tau$, which is empirically set to 0.05, and $sim(h_1,h_2)$ as the cosine similarity $\frac{h_1 \cdot h_2}{|| h_1|| \cdot || h_2||}$.

\subsection{Data}
\label{subsec:data}

We use datasets from two different domains to evaluate the domain adaptation abilities of our approach. We randomly split both domain-specific datasets into 90\% training and 10\% test datasets. 

\paragraph{SciDocs}

The SciDocs dataset \cite{cohan-etal-2020-specter} consists of scientific papers and their citation information. As model input, we concatenate the titles and abstracts of papers with the [SEP] token. Since our model has a maximum input length of $512$ tokens, the input is cut off after this threshold. A positive sample is defined as a directly referenced paper for each anchor sample. A negative sample is a paper referenced by the positive sample but not by the anchor sample itself. This approach ensures that all samples address the same topic, but the positive sample is more related to the anchor sample than the negative one.

\paragraph{AskUbuntu}

The AskUbuntu dataset \cite{lei-etal-2016-semi} consists of user posts from the technical forum AskUbuntu. It already includes sentence pairs that are deemed similar. Therefore, anchor- and positive samples are easily found.
Since the dataset inherently consists of sentences about a similar topic, the operating system Ubuntu, negative sentences can easily be retrieved by sampling different sentences. The dataset originates from a technical domain and is quite different from the scientific domain of SciDocs.

\section{Evaluation}
\label{sec:results}

Table \ref{tab:evaluation} shows the results obtained when adapting sentence embedding models to different domains with adapters. To put the adapter results into perspective, we also evaluate the performance of the SimCSE base model, which is not adapted to the specific domains, as a lower bound. Furthermore, we use traditional domain-specific fine-tuning by training all parameters of the SimCSE base model with the respective loss functions as upper bounds. 

The evaluation reveals that adapter-based domain adaptation yields competitive results compared to fine-tuning the entire base model. In particular, the Houlsby and Pfeiffer adapters perform very well with both loss functions, even though they use only a fraction of the parameters of the upper bounds. The slightly larger K-Adapter, however, performs considerably worse than the other adapters investigated. We conclude that the bottleneck architecture is more suitable than the external plug-in architecture for domain adaptation of sentence embedding models. In particular, the Houlsby adapter, although the smallest among the adapters investigated, yields the best results for both loss functions. Using the out-of-the-box SimCSE model without domain adaptation results in considerably worse performance, indicating the overall importance of domain-specific fine-tuning for sentence embedding models. 

Furthermore, the contrastive loss function $\ell_{2}$ performs consistently better than $\ell_{1}$. Our results align with the observations of \citet{gao-etal-2021-simcse} who conclude that the contrastive objective ensures a distribution of embeddings around the entire embedding space. In contrast, $\ell_{1}$ may yield learned representations occupying a narrow vector space cone, which severely limits their expressiveness. 

From the obtained results, we conclude that using the Houlsby-Adapter architecture together with the contrastive objective $\ell_{2}$ is most suitable for parameter-efficient domain adaptation of sentence embedding models. This adapter approach shows performance that is within 1\% of the supervised, entirely fine-tuned SimCSE model, while only training approximately 3.6\% of the parameters.

\section{Conclusion}
\label{sec:conclusion}

In this work, we proposed the use of adapters for parameter-efficient domain adaptation of sentence embedding models. In contrast to fine-tuning the entire sentence embedding model for a particular domain, adapters add a small number of new parameters that are updated during training while the weights of the sentence embedding model are fixed. We showed that adapter-based domain adaptation of sentence embedding models yields competitive results compared to fine-tuning the entire model, although only a fraction of the parameters are trained. In particular, we show that using the Houlsby-Adapter architecture together with a contrastive objective yields promising results for parameter-efficient domain adaptation of sentence embedding models.


\bibliographystyle{acl_natbib}
\bibliography{anthology,custom}


\end{document}